\documentclass[conference]{IEEEtran}
\IEEEoverridecommandlockouts
\usepackage{cite}
\usepackage{amsmath,amssymb,amsfonts}
\usepackage{algorithmic}
\usepackage{graphicx}
\usepackage{textcomp}
\usepackage{xcolor, soul}
\usepackage{acronym}
\usepackage{subcaption}
\usepackage{multirow} 
\usepackage{array} 
\usepackage{csquotes}

\def\BibTeX{{\rm B\kern-.05em{\sc i\kern-.025em b}\kern-.08em
    T\kern-.1667em\lower.7ex\hbox{E}\kern-.125emX}}
\begin{document}

\title{FloGAN: Scenario-Based Urban Mobility Flow Generation via Conditional GANs and Dynamic Region Decoupling}

\author{
	\IEEEauthorblockN{Seanglidet Yean\IEEEauthorrefmark{1}, Jiazu Zhou\IEEEauthorrefmark{1}, Bu-Sung Lee\IEEEauthorrefmark{2}, Markus Schläpfer\IEEEauthorrefmark{3}}
\IEEEauthorblockA{\IEEEauthorrefmark{1}\textit{Future Cities Laboratory Global}, 
\textit{Singapore-ETH Centre}, 
Singapore, Singapore\\
Email: seanglidet.yean@sec.ethz.ch, jiazu.zhou@sec.ethz.ch}
\IEEEauthorblockA{\IEEEauthorrefmark{2}\textit{College of Computing and Data Science}, 
	\textit{Nanyang Technological University}, 
	Singapore, Singapore\\
	Email: ebslee@ntu.edu.sg}
\IEEEauthorblockA{\IEEEauthorrefmark{3}\textit{Department of Civil Engineering and Engineering Mechanics}, 
	\textit{Columbia University}, 
	New York, USA\\
	Email: m.schlaepfer@columbia.edu}
}
\maketitle

\begin{abstract}
The mobility patterns of people in cities evolve alongside changes in land use and population. This makes it crucial for urban planners to simulate and analyze human mobility patterns for purposes such as transportation optimization and sustainable urban development.  Existing generative models borrowed from machine learning rely heavily on historical trajectories and often overlook evolving factors like  changes in population density and land use. Mechanistic approaches incorporate population density and facility distribution but assume static scenarios, limiting their utility for future projections where historical data for calibration is unavailable. This study introduces a novel, data-driven approach for generating origin-destination mobility flows tailored to simulated urban scenarios. Our method leverages adaptive factors such as dynamic region sizes and land use archetypes, and it utilizes conditional generative adversarial networks (cGANs) to blend historical data with these adaptive parameters. The approach facilitates rapid mobility flow generation with adjustable spatial granularity based on regions of interest, without requiring extensive calibration data or complex behavior modeling. The promising performance of our approach is demonstrated by its application to mobile phone data from Singapore, and by its comparison with existing methods.   
\end{abstract}

\section{Introduction}
As land use and population densities evolve, urban mobility patterns undergo significant transformation. For effective urban infrastructure planning, ranging from transportation network design~\cite{Alessandretti2023} and traffic flow optimization~\cite{Caceres2021} to sustainable urban development~\cite{YANG2025106292}, understanding both current mobility patterns and future scenarios is essential. Analyzing projected mobility flows enables planners to anticipate future urban dynamics and assess the impact of proposed interventions.

Existing methods for mobility flow generation or simulation have primarily relied on models such as the gravity model \cite{barbosa2018human}, the radiation model \cite{simini2012universal}, a recently introduced visitation law \cite{schlapfer2021universal}, or the exploration and preferential return (EPR) model \cite{pappalardo2016human}, differing in their assumptions about human behavior and scale. While the gravity model, as well as the \enquote*{parameter-free} radiation and visitation models, focus on aggregate factors like distance and population, the EPR (and its extensions such as the TimeGeo framework \cite{timegeo}) incorporate individual decision-making based on preferences; i.e. the desire to revisit or explore new locations. These methods require either parameter calibrations on historic data (gravity and EPR-based models), which limits their application for future scenarios, they may have limited predictive capabilities at fine-grained intra-urban scales (radiation model), or may not sufficiently consider location-specific factors such as accessibility (visitation law).

Recently, reinforcement learning (RL) and generative adversarial networks (GANs) have emerged as promising techniques for mobility simulation, offering enhanced flexibility and predictive accuracy. 
RL has been used to model decision-making for mobility generation, where it models intelligent agent behavior and optimal paths in complex urban systems \cite{GLASS20223367}. However, it highlights the high computational demands and challenges in producing reliable, realistic data, which is especially important when data is sparse or highly variable. Among these, GANs hold particular promise, as they can generate diverse, high-quality synthetic mobility data with less reliance on extensive real-world interaction and computational resources. In this paper, we explore the application of GANs for future mobility projections, aiming to enhance the reliability and scalability of urban mobility modeling and simulation.

In the era of big data and data-driven methodologies, generative adversarial networks (GANs) have been utilised to simulate complex systems, including urban mobility, by generating synthetic data that closely mimics real-world patterns. In the context of mobility generation, existing methods like CollaGAN \cite{LEE2025105125} have successfully integrated multiple data sources such as Household Travel Survey (HTS) and Transit Smart Card (SC) data to generate realistic activity schedules. However, these approaches are highly dependent on large, complex datasets, which can be difficult to obtain in data-scarce environments. In addition, other existing methods like MoGAN \cite{mogan} and TrajGDM \cite{trajgdm}, which focus on historical mobility trajectories, often struggle to account for changing factors such as land use and population density shifts.
 
To address these challenges, this study proposes a novel GAN-based approach towards generating projected, scenario-based origin-destination (OD) mobility flows for a simulated urban environment. We intend to challenge the conventional application of GANs, which have seen great success in tasks such as RGB image generation with style changes, by applying them to mobility and OD flow generation. Our proposed method leverages GANs to fuse historical mobility data, such as GPS-based mobile phone data, with style-controlling options, such as base-map configurations, allowing for scalable and adaptable mobility simulations. This approach provides a solution for environments where multi-source, extensive data is scarce or unavailable, which enables accurate prediction of future urban dynamics and evaluation of proposed interventions without relying on a large amount of historical data.

The key contributions of this paper are: 
\begin{itemize}
    \item A data-driven approach to mobility flow generation that (1) requires a relatively small amount of calibration data, (2) operates independently of intricate behavior models requiring expert input, and (3) enables rapid data generation within seconds using a pre-trained model.
    \item A flow decoupling technique leveraging dynamic mapping to enhance the flexibility of GANs for mobility simulations across various regions or specific areas of interest.
    \item Enhanced user control over the generated data, ensuring adaptability to diverse planning and simulation needs.
\end{itemize}

\section{Methodology}

\begin{figure*}[h!]
    \centering
    \includegraphics[width=0.75\linewidth]{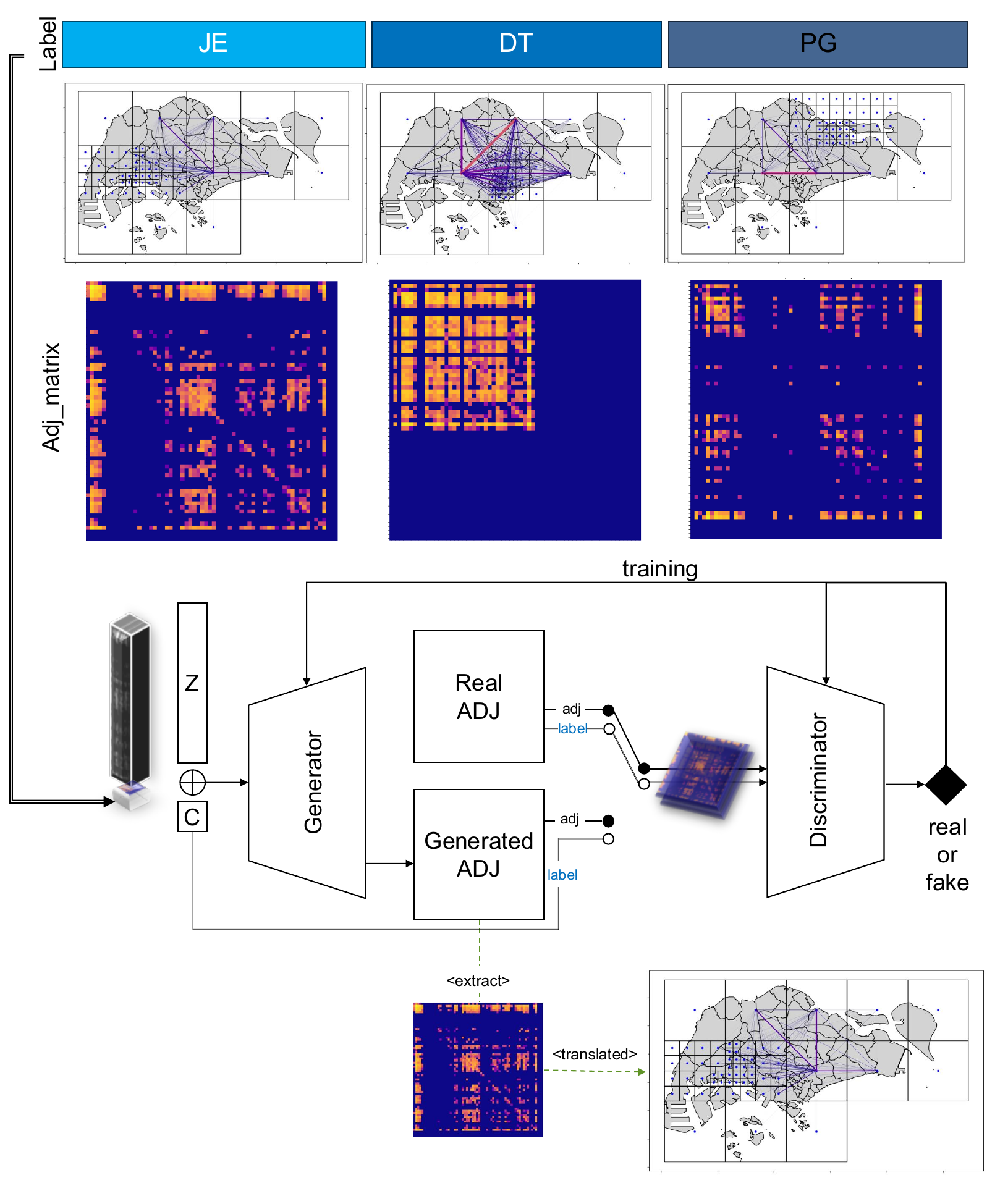}
    \caption{Overview of the proposed FloGAN architecture.}
    \label{fig:flogan}
    \centering
\end{figure*}

\subsection{FloGANs: Architecture Design}

To achieve the mobility OD flow generation with added conditions, we design a methodology, termed \enquote*{FloGAN}, consisting of three main components (Fig.~\ref{fig:flogan}): 
\begin{itemize}
    \item \textbf{Encoder/Decoder}: This component performs representation transformation by converting individual-based trajectory data (GPS traces from mobile phones) into aggregated and structured OD flows. The transformation leverages a dynamic map representation as the spatial basis. The resulting OD flows are subsequently encoded as adjacency matrices to enable further analysis and learning. This approach overcomes the limitations that $i$) GANs typically require a static input size (often requiring a 64$\times$64 image), and $ii$) that an image is usually split into equally sized grid cells, which causes a loss in  flexibility for specific region-to-region mobility flows. 
    \item \textbf{Conditional Encoding}: This component introduces auxiliary information to guide the generative process, enabling control over the \enquote*{style} or characteristics of the generated OD flow data (e.g., time of day, land use scenario).    
    \item \textbf{Generative Model}: This component represents the chosen GAN architecture.
\end{itemize}

\subsection{Encoder/Decoder: Representation Transformation with Dynamic Region-Decoupling}

\begin{figure}[h]
    \centering
    \includegraphics[width=\linewidth]{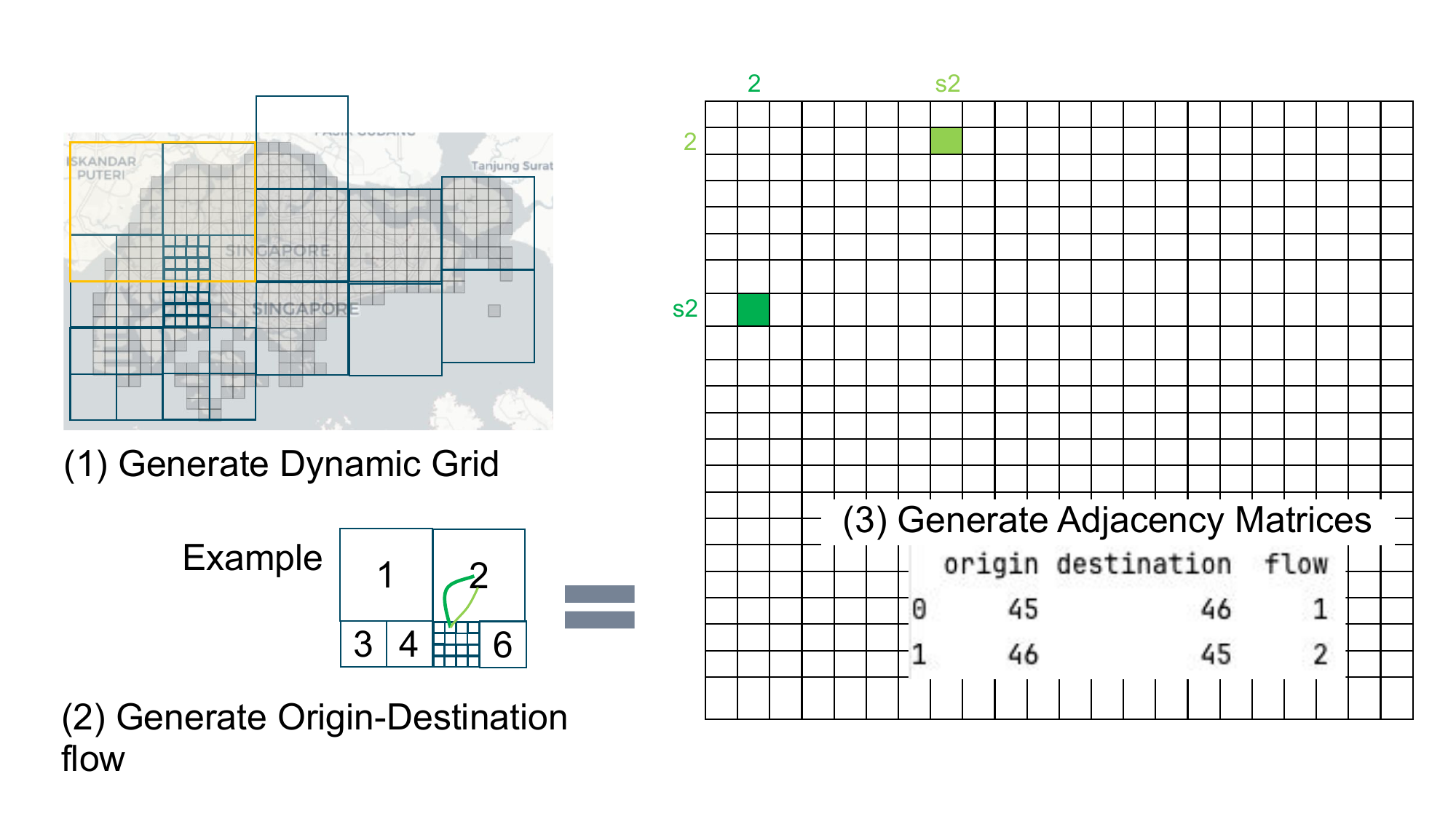}
    \caption{Exemplary representation transformation with dynamic region decoupling. The grid cells can be adjusted based on user-defined regions of interest.}
    \label{fig:dymmap}
    \centering
    \vspace{-0.3cm}
\end{figure}

To address the trade-off between spatial resolution and computational efficiency, we introduce a dynamic map (see Fig.~\ref{fig:dymmap}) that adaptively defines focus regions. The goal is to capture fine-grained mobility patterns in specific areas of interest while maintaining a coarser resolution elsewhere. This approach avoids training the model on the entire map at full resolution, which would be inefficient and prone to overfitting.

Unlike static gridding approaches, the dynamic map enables selective spatial attention without violating the continuity of mobility flows across regions. Since OD flows inherently involve connections between distant zones, naive spatial partitioning is unsuitable. The dynamic map offers a solution by enabling localized focus while maintaining global consistency and abiding by the restricted resolution of the GAN models.

To align with the input requirements of our generative model, the OD flow is conventionally represented as a weighted adjacency matrix $A_{n \times n}$, where $n \le 64$ is the number of spatial grid cells. Each entry $a_{i,j}$ denotes the number of individuals traveling from location $i$ to location $j$. If the number of active grids is smaller than the model’s required input size, the matrix is zero-padded accordingly.
\begin{figure}[h!]
    \centering

    \begin{subfigure}[b]{\linewidth}
        \includegraphics[width=\linewidth]{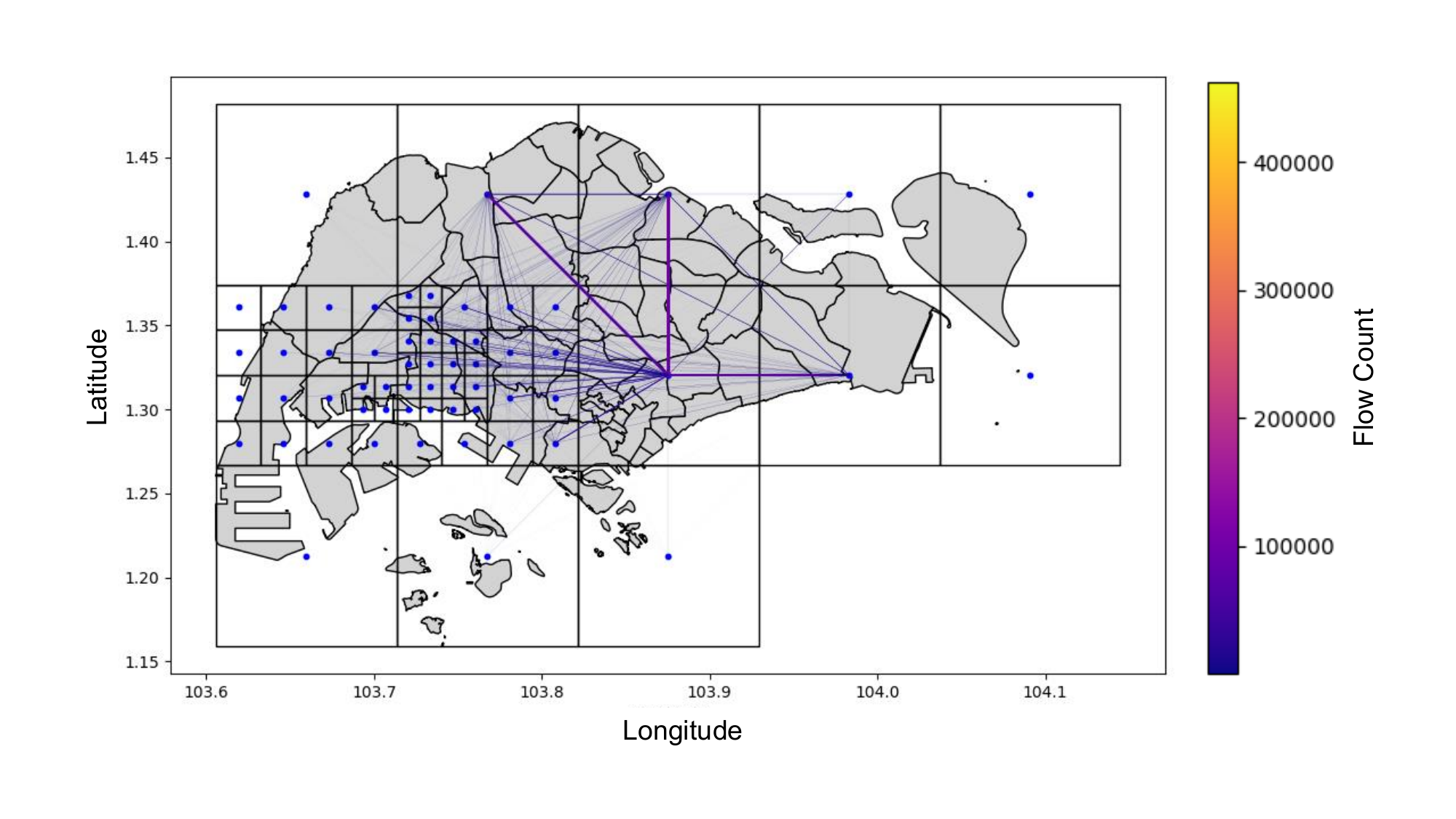}
        \caption{Dynamic map and resulting OD flow.}
        \label{fig:od2adj_a}
    \end{subfigure}
    \hfill
    
    \begin{subfigure}[b]{0.8\linewidth}
        \includegraphics[width=\linewidth]{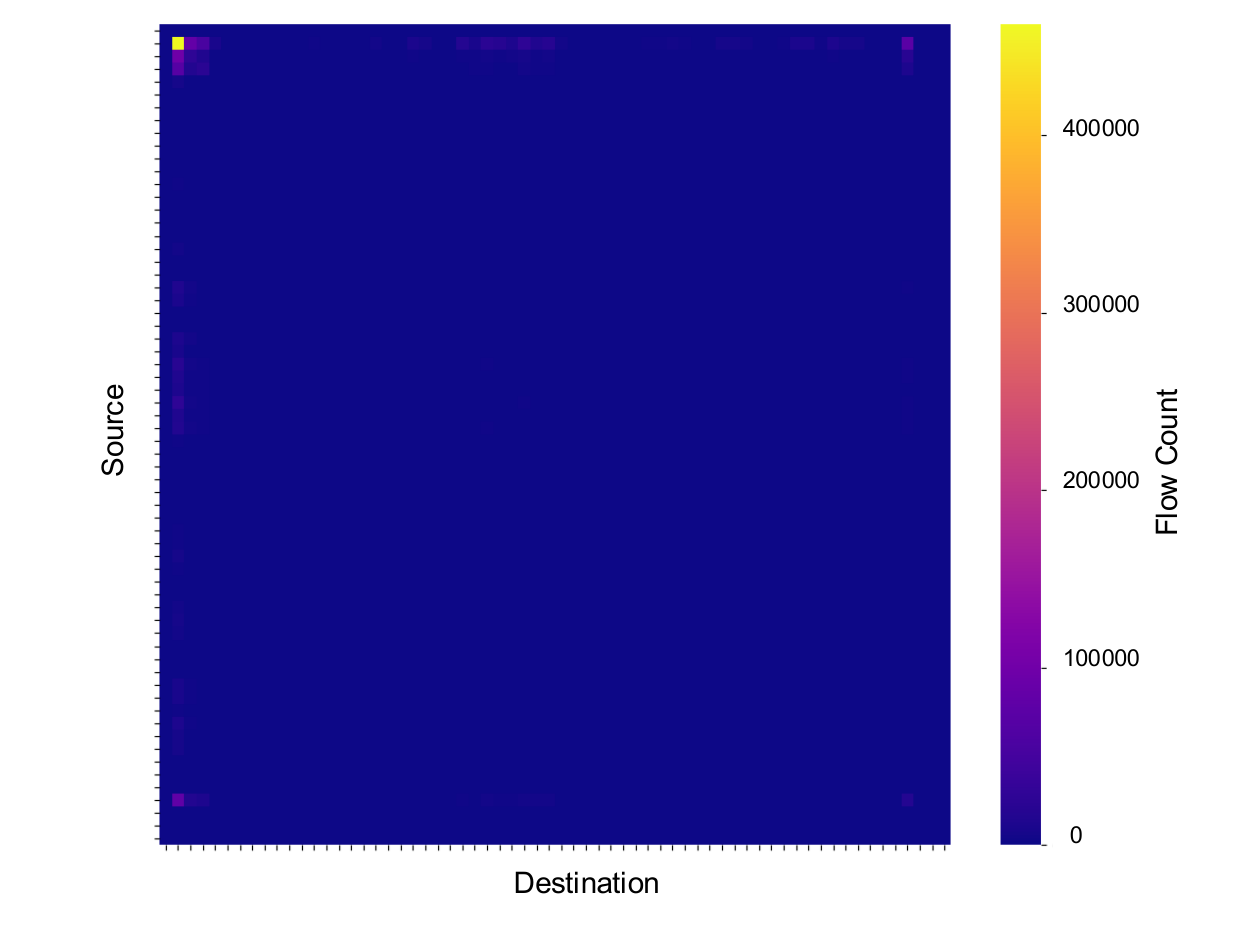}
        \caption{Adjacency Matrix (without normalisation)}
        \label{fig:od2adj_b}
    \end{subfigure}
    \hfill
    
    \begin{subfigure}[b]{0.8\linewidth}
        \includegraphics[width=\linewidth]{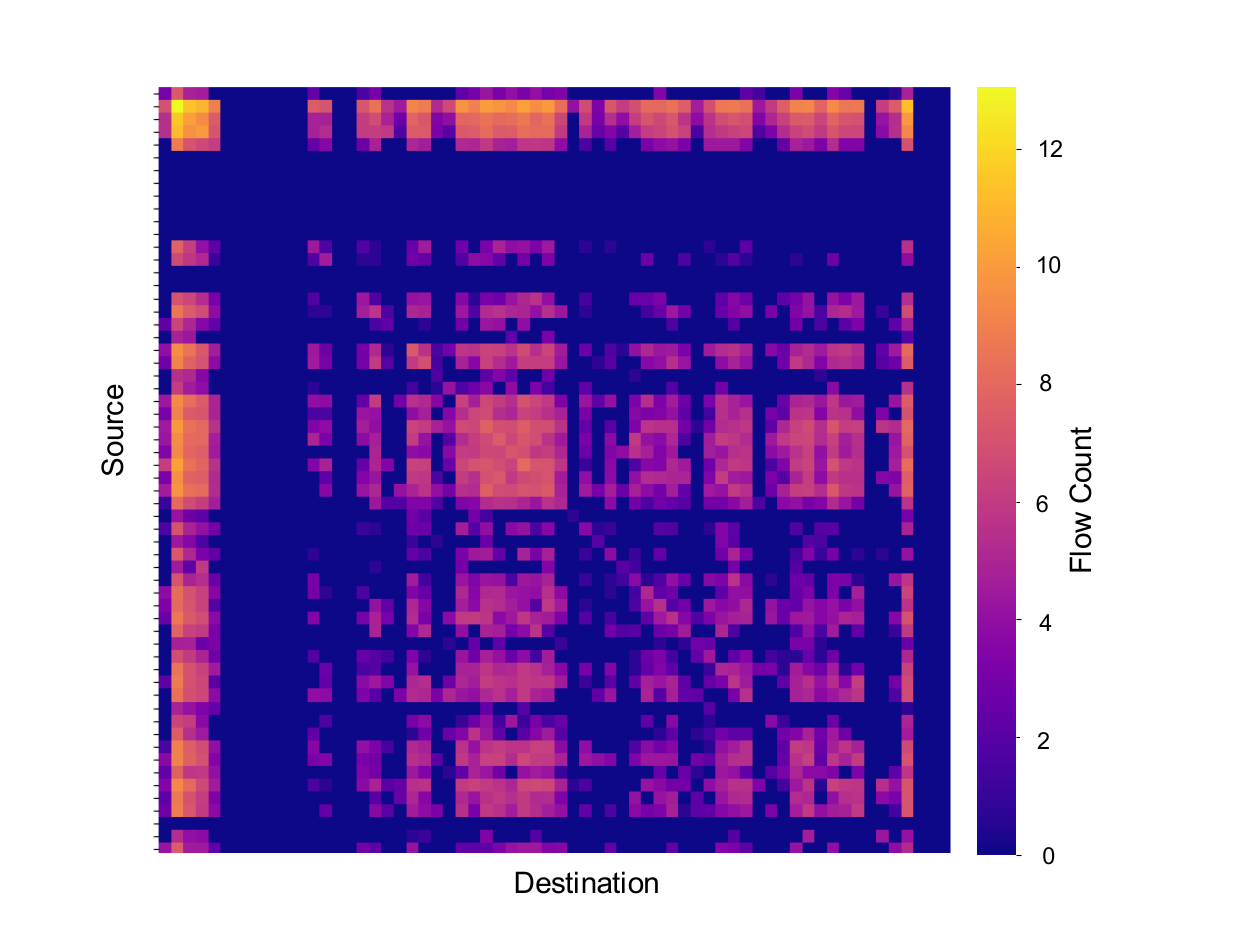}
        \caption{Adjacency Matrix (with logarithm normalisation)}
        \label{fig:od2adj_c}
    \end{subfigure}
    \caption{Data representation (adjacency matrix) of the origin-destination (OD) flows with a dynamic grid. Each subfigure shows a different processing step or configuration.}
    \label{fig:od2adj}
    \vspace{-0.5cm}

\end{figure}

In this study, we use three different sizes of grid cells: 1.5$\times$1.5, 3$\times$3, and 12$\times$12 km$^2$, based on empirical testing under the constraint of a maximum total number of grid cells $n_{\max }=\,\,$64, aligned with GAN input dimensions (64$\times$64). The region of interest is subdivided into finer 1.5$\times$1.5 km$^2$ cells to capture detailed patterns, while the surrounding area uses 3$\times$3 km$^2$ cells, nested within a broader 12$\times$12 km$^2$ spatial extent. 

The OD flow is then aggregated from the trajectory data to represent the trip volumes from one grid cell to another. This aggregated flow is then encoded as a weighted adjacency matrix, where each entry represents the number of trips from one cell to another.

The adjacency matrix proposed in \cite{mogan} did not consider normalisation. However, mobility flow data often exhibit large variability with highly skewed distributions, where small changes in flow volumes are relatively insignificant compared to the broader range. To address this issue, we apply a natural logarithmic transformation to the OD flow values, effectively compressing the data range and reducing skewness:
\begin{equation}
\text{Norm}(x) = \log_e(1 + x),
\end{equation}
where $x$ represents the values in the adjacency matrix. We use $\log_e(1+x)$ instead of $\log_e(x)$ in order to stabilise numerically when the value of $x$ is small. Figure~\ref{fig:od2adj} illustrates the process of data representation generation using the dynamic map and the derived OD flow. As a result of the logarithmic transformation, the adjacency matrix shown in Fig.~\ref{fig:od2adj}c emphasizes a greater number of OD flows, since the visibility of lower-volume connections in the pixel-based representation is enhanced.

\subsection{cGANs Model Configurations}

The core architectural design of the proposed methodology centers on selecting a base generative framework that supports the integration of auxiliary information. This enables conditional control over the \enquote*{style} or characteristics of the generated data. Once an appropriate base architecture is identified, it can be adapted to align with our specific data representation and generation requirements.

The method in \cite{mogan} is based on a deep convolutional GAN (DCGAN) \cite{dcgan} that was built to allow for images as the training input, and it consists of a convolutional block in the generative and discriminative networks. It works well when we transform our inputs into an adjacency matrix where the DCGAN model observes this input as a pixel-based image. The TrajGDN method \cite{trajgdm} uses a slightly different approach where the input representation was transformed using a trainable encoder and fed into a diffuser network. While TrajGDN is good for interpretability and training stability, it is slower in inference as compared to GANs-based methods, and it does not provide flexibility in adding auxiliary information.

Among various generative approaches, conditional GANs (cGANs) are particularly well-suited for mobility flow generation due to their ability to model complex data distributions while allowing direct conditioning on auxiliary variables. This provides fine-grained control over the generation process, which is critical for capturing the spatiotemporal dependencies inherent in mobility data. In contrast, variational autoencoders (VAEs) typically encode conditioning information into the latent space, making it harder to enforce precise control over outputs. Diffusion models, while offering stability and strong generation quality, involve computationally intensive iterative sampling procedures. cGANs offer a more efficient, single-pass generation process, with greater flexibility in handling structured input formats like adjacency matrices.

\begin{figure*}[h!]
    \centering
        \includegraphics[width=0.95\linewidth]{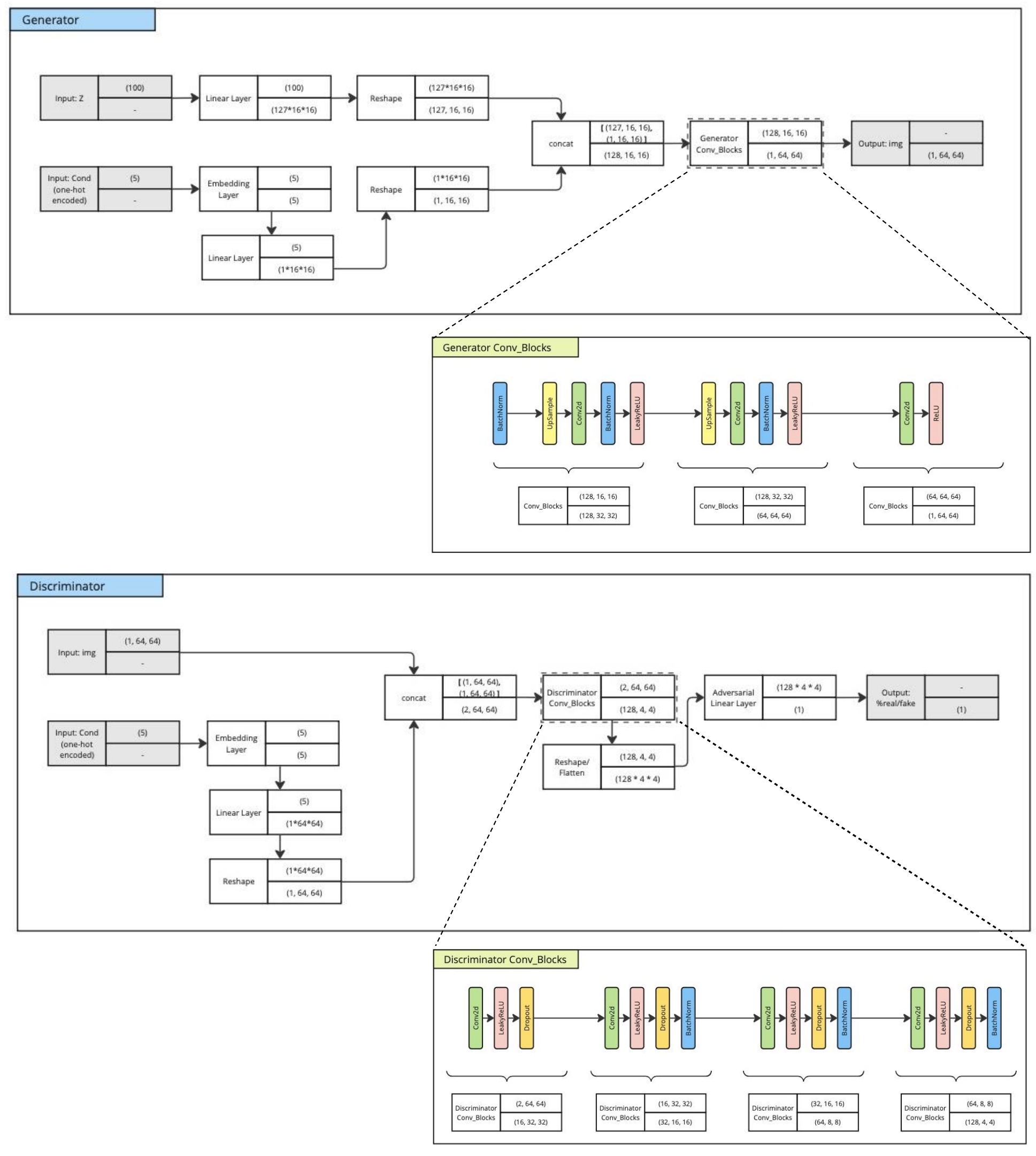}
        \label{fig:cgan-generator}
    \caption{FloGAN: generator and discriminator of the conditional GAN (cGAN) architecture.} 
    \label{fig:cgan-architecture}
    \vspace{-0.2cm}
\end{figure*}

In our proposed method, we extend the DCGAN approach in \cite{mogan} by adopting a conditional GAN (cGAN) architecture, as shown in Fig. \ref{fig:cgan-architecture}. The conditional component allows for the flexible integration of auxiliary information, enhancing the model’s adaptability to diverse urban scenarios. This auxiliary information can be incorporated either as an encoded representation or as additional adjacency matrix-like input. 
Training a separate GAN for each dynamic map would result in an inefficient use of the data, reduce the chance to learn shared structural patterns, and significantly increase computational overhead. In contrast, the cGAN facilitates scalable and data-efficient learning by sharing parameters across map types and specializing outputs through auxiliary inputs; hence, improving both performance and adaptability.

In the generator architecture, the 1D conditional input is first passed through a trainable embedding layer, followed by a linear transformation that projects it to a spatial feature map of dimension $1 \times 4 \times 4$. The adjacency matrix, representing the origin-destination (OD) flow information, is also processed through a linear layer to reshape it into a $127 \times 4 \times 4$ (without conditional input, the adjacency flow matrix would be enlarged to $128 \times 4 \times 4$).
The reshaped adjacency matrix and the conditional feature map are then concatenated along the channel dimension, forming a unified $128 \times 4 \times 4$ input. This input is subsequently passed through a series of convolutional layers, producing a $1 \times 64 \times 64$ output image, which corresponds to the generated adjacency flow map.

The discriminator architecture mirrors this process in reverse. Given a $1 \times 64 \times 64$ input image, either a real or generated adjacency matrix, the same 1D conditional vector is passed through an embedding layer and a linear transformation to produce a $1 \times 64 \times 64$ input. This input is concatenated with the input image along the channel dimension to form a $2 \times 64 \times 64$ input. This combined input is then passed through several convolutional layers followed by linear layers; thus, producing a scalar output. A sigmoid activation function is applied to this scalar, yielding a value in the range $[0, 1]$ that represents the discriminator’s confidence if the input image is real or fake.

\section{Experimental Results and Analysis}
\subsection{Experimental Setup}
\subsubsection{Dataset}
The mobility data used in this study was provided by CITYDATA.ai \cite{citydata2022}, consisting of anonymized mobile phone records in Singapore during a 30-day period from September 1 to September 30, 2019. We performed several preprocessing steps as described in \cite{zhou2024estimating}, resulting in the trajectories (i.e., time-stamped sequences of visits to different locations) of 104,976 active users.

For this study, the mobility records were then aggregated into OD flows based on spatial and temporal groupings \cite{barbosa2018human}. The spatial aggregation was defined through six dynamic maps, each of which sets a focus on a different urban region in Singapore: Jurong East (JE), Downtown (DT), Punggol (PG), and three subzones within the Tampines region (TM0, TM1, TM2). We use the name of these dynamic maps as the conditional input. Temporally, the data was segmented into six distinct time groups per day to capture different mobility patterns throughout the day, based on the reported travel patterns of Singapore \cite{zeng2017visualizing}:
\begin{itemize}
    \item Group1: Morning Travel (5:00 AM – 8:00 AM)
    \item Group2: Work Morning (8:00 AM – 11:00 AM)
    \item Group3: Lunch Travel (11:00 AM – 2:00 PM)
    \item Group4: Work Afternoon (2:00 PM – 5:00 PM)
    \item Group5: Dinner Travel (5:00 PM – 8:00 PM)
    \item Group6: Night (8:00 PM – 5:00 AM)
\end{itemize}

Each day thus contributes six OD matrices (one for each time group), and each OD matrix corresponds to one of six dynamic spatial configurations (i.e., dynamic maps). These OD flows were further transformed into adjacency matrices \cite{barbosa2018human}.
In total, the dataset used in this experiment comprises 1,080 data points (30 days, 6 time groups, 6 dynamic maps).

\subsubsection{Experimental Study Design}
To evaluate the performance of our proposed model, FloGAN, we compare it against the following baseline methods:
\begin{itemize}
    \item MoGAN \cite{mogan}: A generative adversarial network designed for synthetic mobility flow generation. It has been previously applied to datasets from New York City and Chicago, focusing on bike and taxi mobility flows.
    \item Gravity model \cite{barbosa2018human}: A traditional and widely-used approach in mobility modeling, based on the analogy of gravitational force. The flow magnitude between two regions is assumed to be proportional to the product of their masses (e.g., population or activity levels) and to decrease with increasing distance between them (e.g., as a power law function).
\end{itemize}

To evaluate the quality of the generated mobility flows, we adopt a combination of statistical, image-based, and domain-specific metrics. The checksum metric serves as a basic validation tool by ensuring that the generated OD flow values fall within a plausible range, providing an early indication of the model’s stability and correctness. Given that each OD matrix can be represented as a grayscale image, we apply standard image quality metrics commonly used in GAN evaluation, including Structural Similarity Index Measure (SSIM), Peak Signal-to-Noise Ratio (PSNR), and Fréchet Inception Distance (FID). These metrics assess the visual similarity, pixel-level accuracy, and distributional realism between the generated and real OD matrices. In addition, we use the Common Parts of Commuters (CPC), which quantifies the similarity between the generated number of trips between each OD pair and the real data and is a widely used metric in the mobility modeling domain~\cite{barbosa2018human}. The coefficient takes values between 0, when no agreement is found, and 1, when there is a perfect match between the model and the data.

\subsection{Checksums Test}

\begin{table}[b!]
\caption{Checksum of experiment data}
\label{tab:checksum}
\centering
\begin{tabular}{|l|l|l|}
\hline
\textbf{Experiment Data} & \textbf{Maximum (MAX)} & \textbf{Average (AVG)} \\ \hline
Ground Truth             & 84,480           & 29.28            \\ \hline
FloGANs                  & 442,412.39       & 28.62            \\ \hline
MOGAN                    & 1,202,603.28     & 54.76            \\ \hline
Gravity                  & 240,095.27       & 13.62            \\ \hline
\end{tabular}
\end{table}

MAX and AVG of Table~\ref{tab:checksum} show the most extreme and average value of the generated OD flow, respectively. Examining the MAX values shows that all mobility generators tend to overestimate the extreme values when compared to the ground-truth data. The overestimation could be due to the difficulty in capturing such extreme events, often not well-represented by historical data.
Among all, the gravity model produces the closest match, with an overestimation factor of 2.84. 
For the AVG value, we observe that FloGAN produces the closest flow average when compared to other methods. This indicates that FloGAN may has the capability to capture the mobility flow distribution of the ground truth.

\subsection{Evaluation - Image Quality}
\begin{figure*}[h!]
    \centering
    \includegraphics[width=\linewidth]{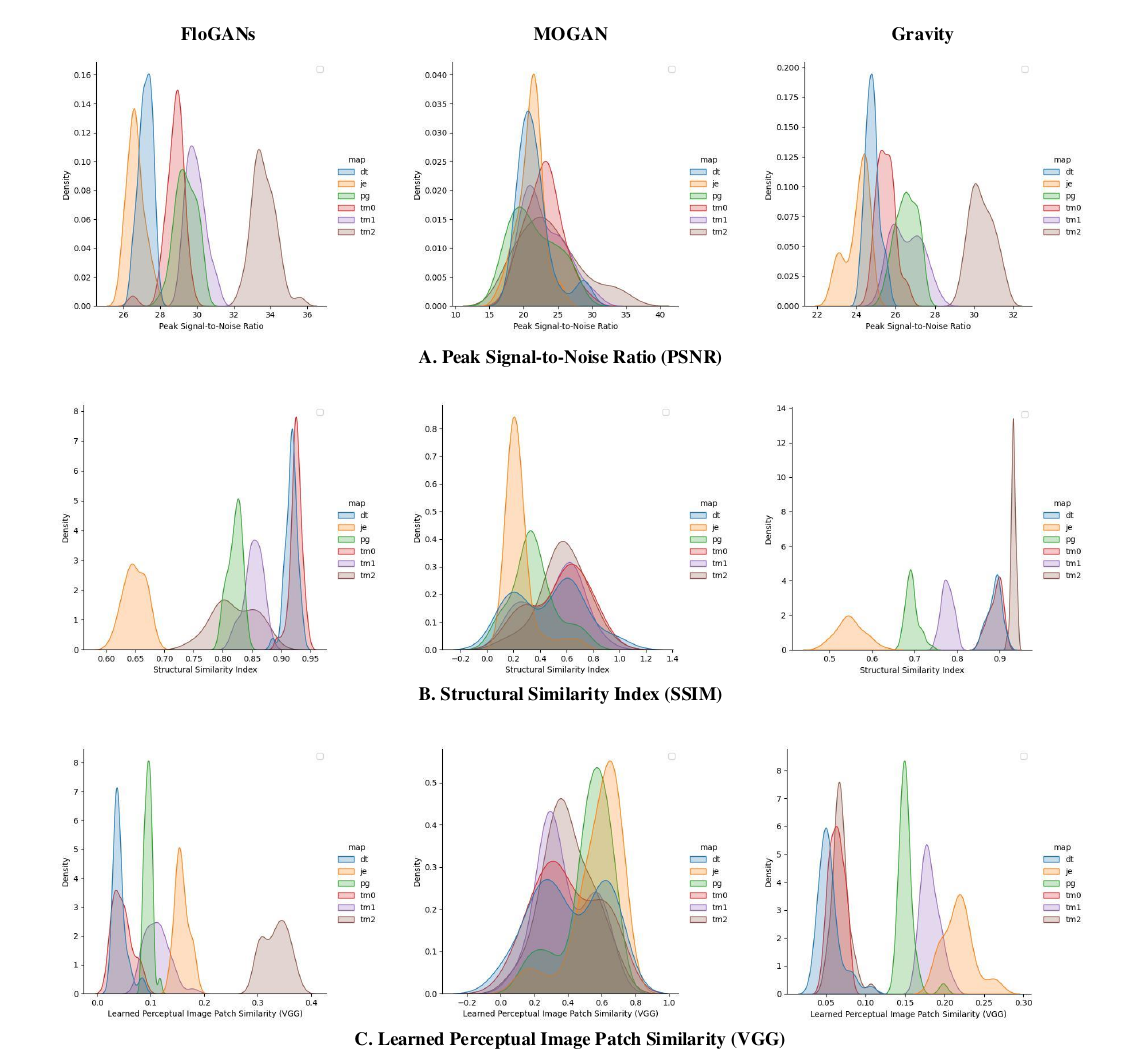}
    \caption{Image-based evaluation of the generated data (kernel density plots). A. Peak Signal-to-Noise Ratio (in dB), with higher values corresponding to higher pixel-level similarity. B. Structural Similarity Index. Values are in the range [-1,1], with values closer to 1 corresponding to higher structural similarity. Note that the kernel density plot smoothens over the distributions, which may lead to estimates outside the theoretical range. C. Learned Perceptual Image Patch Similarity. Values are in the range [0,1], with values closer to 0 corresponding to higher perceptual similarity.}
    \label{fig:eval_img}
    \centering
\end{figure*}

PSNR (Fig.~\ref{fig:eval_img}.A) quantifies the pixel-wise similarity of the generated and real images. As such, it shows the differences in mobility flow magnitude for each OD pair. The results are grouped per dynamic map. FloGAN and the gravity model show different performances with respect to the dynamic maps, while MoGAN generates the results for all dynamic maps within the same range. That is expected because MoGAN is not using a conditional GAN and thus lacks control over the style or context of the data it generates. As such, it is trained on the combined set of adjacency matrices from all dynamic maps without any conditional input to indicate which map each sample belongs to and, as a result, it produces outputs that average over the entire training distribution.

The gravity model is able to show the differences, since it is based on a random training sample of the selected dynamic map. In other words, the gravity model can be controlled by the dynamic map setting. The range of PSNR results of FloGAN and MoGAN shows that FloGAN has a higher performance as its minimum value is higher than MoGAN’s minimum. FloGAN and the gravity model show comparable results with regards to the performance per dynamic map. 

SSIM (Fig.~\ref{fig:eval_img}.B)  focuses on structural similarity rather than pixel-based similarity. It focuses on contrasting aspects of the generated data compared to the real data (aligning more with human visual perception, if we were to eyeball the results). This evaluation focuses on the extreme data being generated and how close they are to the real data.
Overall, FloGAN and the gravity model have a better range in their results than MOGAN (0.5 - 0.95). Starting at a minimum value of $\approx\,$0.6, FloGAN shows a better performance compared to the other two models. This indicates that the extreme contrasting data generated by FloGAN align more closely with the ground truth. While FloGAN and the gravity model achieve comparably strong results overall, they show different performances across the dynamic maps. FloGAN outperforms the gravity model for almost all dynamic maps, demonstrating superior adaptability and performance consistency. Notably, FloGAN produces better results for Jurong East (je) as it has the most complicated mobility flows due to its composition of land use, being the hub of residential, commercial, and business activities. FloGAN also performs better in the Downtown (dt), where the concentration of commercial and business functions presents a similarly challenging mobility landscape. The gravity model performs only slightly better than FloGAN for Tampines 2 (tm2), which is adjacent to Tampines 0 (tm0) and has similar, residential-oriented land use. 

LPIPS (Fig.~\ref{fig:eval_img}.C) measures the perceptual similarity between image patches using features extracted by a deep convolutional neural network (VGG). Unlike traditional pixel-wise metrics, LPIPS captures high-level semantic similarities that align more closely with human visual perception. The results of MoGAN vary across the full $[0, 1]$ range, suggesting a lack of structure in the generated outputs, likely due to the absence of conditioning mechanisms (dynamic map control). By contrast, FloGAN and the gravity model consistently achieve lower LPIPS values, indicating better perceptual similarity with ground truth flows. However, in the case of Tampines 2 (tm2), FloGAN's performance declines to the 0.3–0.4 range, while the gravity model maintains stronger perceptual alignment. On Jurong East (je), FloGAN outperforms the gravity model.

\subsection{Evaluation - Common Parts Commuters (CPC)}

\begin{table}[t!]
\centering
\caption{Evaluation - Common Parts of Commuters}
\label{tab:eval_cpc}
\begin{tabular}{|l|l|l|l|}
\hline
\textbf{Dynamic map} & \textbf{FloGAN} & \textbf{MoGAN}   & \textbf{Gravity} \\ \hline
JE       & 0.7381  & 0.5260  & 0.2279  \\ \hline
DT       & 0.8139  & 0.2959  & 0.5118  \\ \hline
PG       & 0.9292  & 0.2567  & 0.2824  \\ \hline
TM0      & 0.8741  & 0.4141  & 0.4832  \\ \hline
TM1      & 0.9281  & 0.4398  & 0.4533  \\ \hline
TM2      & 0.8884  & 0.5260  & 0.3957  \\ \hline
\textbf{Average}  & \textbf{0.8619 } & \textbf{0.40975} & \textbf{0.3923}  \\ \hline
\end{tabular}
\end{table}

Overall, as shown in Table~\ref{tab:eval_cpc}, FloGAN achieves the highest performance with an average CPC score of 0.8619 across all dynamic maps, indicating the greatest alignment with the real mobility flows. Regarding individual dynamic maps, FloGAN consistently outperforms the other two models across all maps. In contrast, MoGAN produces results highly variable scores, reflecting its unconditioned generation process. Meanwhile, the lower scores of the gravity model across the different maps suggest its limited adaptability to diverse historical mobility patterns. These results confirm that FloGAN delivers superior performance compared to both MoGAN and the gravity model.

\section{Conclusion}
This paper presents the results of an ongoing exploration and conceptual development of scenario-based projected mobility flows, specifically, origin-destination (OD) flows, within simulated urban environments over future timelines. The proposed initial framework, FloGAN, is able to incorporate various land use archetypes by including dynamic grid maps and auxiliary information into a conditional generative adversarial network (cGAN). The framework generates mobility data with adjustable resolution based on user-defined regions of interest, enhancing computational efficiency. By using higher resolution in dense or significant areas, it accurately captures mobility patterns while reducing resource demands, enabling faster execution for large-scale and real-time applications.

While the results presented here are promising, our approach needs to be further extended to broaden the consideration of detailed urban characteristics, which are potentially important for the flow prediction in future urban scenarios. To that end, a StyleGAN-inspired method seems particularly promising as it introduces the concept of \enquote*{style} transfer for image generation. Built on the WGAN-GP architecture, which offers improved training stability, this model can incorporate more detailed urban scenarios as auxiliary information (e.g., points of interest distribution) encoded as a \enquote*{style} image to guide the generation process. This will make it possible to predict shifts in mobility patterns caused by changes in urban land use, enabling the assessment of urban planning impacts and the evaluation of the existing or planned transport infrastructure under various land use scenarios. Finally, to make the framework applicable for practical urban planning tasks, it will thereby also be important to quantify the uncertainty~\cite{steentoft2024} of both the assumed urban scenarios and the resulting mobility flows.

\bibliographystyle{IEEEtran}
\bibliography{ref}

\end{document}